\documentclass{pinepaper}

\title{UniIntervene: Agentic Intervention for Efficient Real-World Reinforcement Learning}

\author[1]{Haoyuan Deng}
\author[1]{Yitong Gao}
\author[1]{Yudong Lin}
\author[1]{Haichao Liu}
\author[2]{Zhenyu Wu}
\author[1]{Ziwei Wang{\textdagger}}
\affil[1]{Nanyang Technological University}
\affil[2]{Beijing University of Posts and Telecommunications}

\date{\small \textsuperscript{\textdagger}Corresponding author.}

\papervenue{Preprint}
\paperdate{\today}
\projectpage{https://denghaoyuan123.github.io/UniIntervene-project/}
\correspondence{\href{mailto:author@example.edu}{ziwei.wang@ntu.edu.sg}}

\usepackage{graphicx}
\usepackage[dvipsnames]{xcolor}

\usepackage{amsmath}
\usepackage{mathtools}
\usepackage{mathrsfs}
\usepackage{bbm}   

\usepackage{multirow}
\usepackage{booktabs, tabularx, makecell, colortbl, array}
\newcolumntype{Y}{>{\centering\arraybackslash}X}
\usepackage{siunitx}
\usepackage{vcell}

\usepackage{tikz}
\usepackage{wrapfig}
\usepackage{float}
\usepackage{caption}

\usepackage{tcolorbox}
\usepackage{listings}
\usepackage{courier}

\usepackage{xspace}

\newcommand{\method}{\texttt{\textbf{UniIntervene}}\xspace}

\usepackage{algorithm}
\usepackage[noend]{algpseudocode}

\usepackage{url}
\usepackage{soul}
\usepackage{upgreek}
\usepackage{mwe}
\usepackage{calc}
\usepackage{pifont}
\usepackage{marvosym}

\usepackage{hyperref}

\newcommand{\ourcolor}{gray!9}
\definecolor{DarkGreen}{RGB}{0,128,0}
\definecolor{mymagenta}{HTML}{E1208A}

\begin{document}
\makepinetitle

\begin{pineabstract}
Human-in-the-loop reinforcement learning (HiL-RL) has emerged as an effective paradigm for real-world robotic manipulation, enabling online policy improvement with human guidance.
    However, current HiL-RL frameworks remain intervention-intensive, relying on frequent human corrections to redirect the policy out of unproductive exploration, which incurs high labor cost and limits real-world scalability.
    To address this, we propose \method, an agentic intervention model that detects unproductive exploration and autonomously recovers the policy toward high-value states, taking over the bulk of interventions from human operators. 
    Specifically, \method first performs future-conditioned action-value estimation, predicting the latent consequence of the current action and evaluating its induced value, which provides a more stable progress signal.
    Building on this, a temporal value-risk critic aggregates recent value dynamics and triggers intervention when the estimated value exhibits sustained stagnation or degradation.
    When intervention is required, \method retrieves a high-value recovery target from a memory of past intervention episodes and produces executable corrective actions through a goal-conditioned recovery policy.
    In this way, \method turns intervention from passive human correction into a value-aware recovery process for efficient real-world RL.
    Extensive experiments on diverse real-world manipulation tasks demonstrate that \method improves the average success rate by 8.6\% while reducing human interventions by 57\% relative to state-of-the-art HiL-RL baselines.
\end{pineabstract}

\section{Introduction}


Robotic manipulation is essential for robots in industrial assembly, household service, and unstructured real-world environments~\cite{zhang2024empowering, billard2019trends}.
Recent imitation learning advances have produced general manipulation policies trained from large-scale demonstrations~\cite{kim2025openvla, black2024pi_0}, but offline-trained policies still struggle with out-of-distribution states, contact-rich dynamics, and failure recovery behaviors rarely covered by demonstrations~\cite{chi2025diffusion, deng2025survey}.
Real-world reinforcement learning (RL) addresses these gaps by improving policies through online interaction, yet remains constrained by sparse rewards, unsafe exploration, and low sample efficiency~\cite{zhang2021reinforcement, luo2024serl, chen2024rlingua}.
Human-in-the-loop reinforcement learning (HiL-RL) mitigates these issues by injecting online human intervention and corrective guidance into the learning loop~\cite{luo2024precise, chen2025conrft, xu2025compliant}.
However, current HiL-RL frameworks remain intervention-intensive: the human operator is repeatedly required to redirect the policy out of unproductive exploration before convergence, which incurs substantial labor cost and limits real-world scalability~\cite{luo2024precise, deng2026e2hil}.

Existing efforts to reduce this human burden mainly follow two directions.
The first line improves how intervention data is exploited, reusing takeovers, corrective actions, and recovery trajectories as demonstrations or off-policy supervision to accelerate policy learning~\cite{xu2025compliant, deng2026e2hil, pan2026sop, wang2026learning}.
While effective in improving sample efficiency per intervention, these methods leave the underlying online rollout process unchanged, so the policy still spends substantial interactions in low-value states before a human is solicited.
The second line reduces manual assistance through more autonomous exploration, including reset-free learning, recovery policies, safety critics, and data-efficient policy optimization~\cite{thananjeyan2021recovery, gupta2021reset, su2026ig, gigabrainteam2026gigabrain05m}.
These methods improve training safety and continuity, but remain primarily defensive, avoiding irreversible failures or restoring resettable states without reasoning about whether ongoing exploration is making task progress.
Consequently, neither line addresses the dominant source of human cost in HiL-RL: rollouts that are not unsafe yet fail to progress, leaving human takeover as the only resort. What is missing is an intervention mechanism that detects unproductive exploration online and autonomously recovers the policy toward high-value states, reserving human supervision for the residual cases.

\begin{figure}[t]
  \centering
  \includegraphics[width=1.0\linewidth]{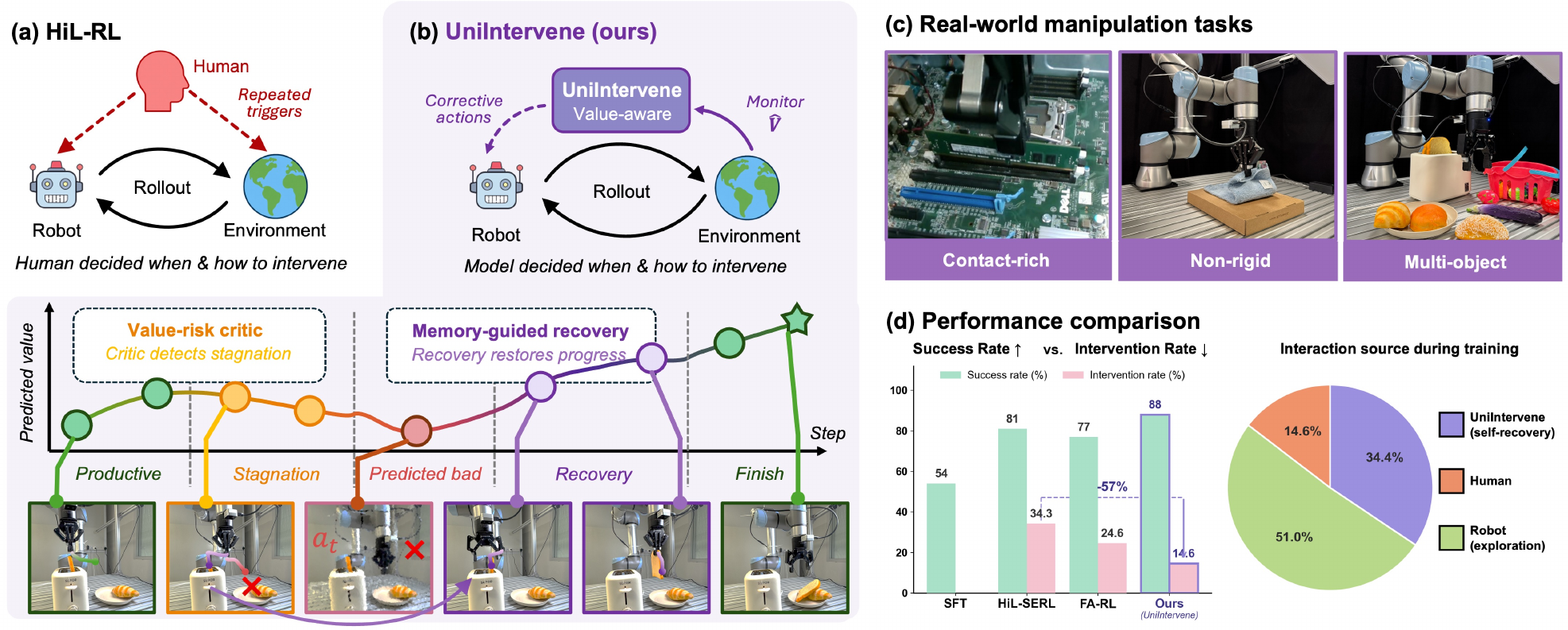}
  \caption{\textbf{\method} replaces human intervention in HiL-RL with a value-aware critic that detects unproductive rollouts and a recovery policy that restores progress, yielding higher success with fewer human interventions on real-world manipulation.}
  \label{fig:teaser}
  \vspace{-10pt}
\end{figure}

To address this gap, we propose \method{}, an agentic intervention model that takes over the bulk of online interventions from human operators during HiL-RL training.
Instead of waiting for failures to occur or for a human to step in, \method{} continuously monitors whether the ongoing rollout is making productive task progress, and autonomously redirects the policy when it is not.
At the core of \method{} is a future-conditioned action-value estimator that predicts the latent consequence of the current action and evaluates the value induced by this predicted future state, providing a more stable progress signal under sparse rewards than estimating value directly from the current observation.
Built on this estimator, a temporal value-risk critic aggregates recent value dynamics over a sliding window and flags unproductive exploration in which the estimated value exhibits sustained stagnation or degradation.
Once triggered, \method{} retrieves a high-value recovery target from a memory of past intervention episodes and produces executable corrective actions via a goal-conditioned recovery policy, returning the rollout to a productive state without human takeover.
In this way, \method{} redirects unproductive interactions into recoverable and informative trajectories, turning intervention into an internal recovery process driven by value-aware decisions.
Our contributions are summarized as follows:
\begin{itemize}
\item We formulate online intervention in HiL--RL as a future-conditioned value-risk decision problem, and propose a temporal value-risk critic that detects unproductive exploration from the dynamics of action-conditioned future value.
\item We develop a memory-guided goal-conditioned recovery policy that retrieves high-value targets from past intervention episodes and produces corrective actions, enabling autonomous recovery without human takeover.
\item Extensive real-world experiments on diverse manipulation tasks show that \method{} achieves 8.6\% higher success rate with 57\% fewer human interventions than state-of-the-art HiL-RL baselines.
\end{itemize}


\section{Related Work}

\noindent\textbf{Human-in-the-loop Reinforcement Learning.}
Human-in-the-loop reinforcement learning (HiL-RL) injects human guidance into the online learning loop to reduce unsafe or unproductive exploration, and existing methods fall into two complementary lines.
The first treats human input as a supervision signal, reusing takeovers and corrective transitions through imitation learning~\cite{Jiang2025TRANSIC, kelly2019hg, mandlekar2020human}, replay buffers~\cite{luo2024serl, ross2011reduction}, residual policy heads~\cite{xu2025compliant, ankile2024juicer}, or off-policy RL updates~\cite{chen2025conrft, luo2024precise, hu2023reboot} to improve sample efficiency, yet still rely on the human operator to decide when intervention is required, treating the trigger as an exogenous event rather than a learnable quantity.
The second treats human input as a continuity mechanism, invoking resets, recovery demonstrations, or proactive help-seeking to keep training tractable after failures~\cite{thananjeyan2021recovery, Kim_2022, yu2025armada, sharma2022state, hu2023imitation}; while some explicitly model when assistance should be solicited~\cite{xie2022askhelpproactiveinterventions, hoque2022thriftydagger, liu2024model}, their criterion is typically tied to safety violations or terminal failures, decoupled from a forward-looking estimate of whether the ongoing rollout is making task progress.
\method{} instead unifies trigger and recovery within a single value-aware decision process: it learns a future-conditioned value estimator that turns visual rollouts into a progress-aligned signal, derives an intervention trigger from the temporal dynamics of this signal, and grounds recovery in retrieved high-value targets from experience.

\noindent\textbf{Autonomous Exploration for Real-world RL.}
Beyond human-in-the-loop supervision, a parallel line of work seeks to make real-world RL more self-sufficient by reducing reliance on human resets and unsafe exploration, and can be grouped into reset-free recovery and safety-constrained exploration.
Reset-free recovery methods aim to remove manual resets by learning reversible behaviors~\cite{eysenbach2017leavetracelearningreset}, recovery policies~\cite{li2026failure, lu2020reset}, forward-backward task pairs~\cite{gupta2021reset, xu2020continual}, or self-induced subgoal curricula~\cite{sharma2021autonomous, zhu2020ingredients} that let the robot keep interacting with limited human assistance, yet recovery is typically driven by reachability or reset distance rather than by an explicit estimate of task value, leaving the robot unable to tell apart being resettable from being on a productive path.
Safety-constrained exploration methods instead confine learning to safe regions using safety critics~\cite{thananjeyan2021recovery, srinivasan2020learning, bharadhwaj2020conservative}, constrained policy optimization~\cite{zhang2025safevla, achiam2017constrained, stooke2020responsive}, or preference-aligned action priors~\cite{zhang2024grape, hejna2310contrastive}; while effective at curbing unsafe behaviors, their constraints are defined over state safety, providing no mechanism to detect rollouts that remain feasible yet make no progress.
\method{} targets precisely this blind spot, asking not whether the robot is resettable or the next state is safe, but whether the ongoing rollout is gaining value, and uses the temporal dynamics of a future-conditioned value estimator to detect stagnation and retrieve high-value recovery targets. 
\section{Methodology}
\label{sec:method}
\begin{figure}[t]
  \centering
  \includegraphics[width=1.0\linewidth]{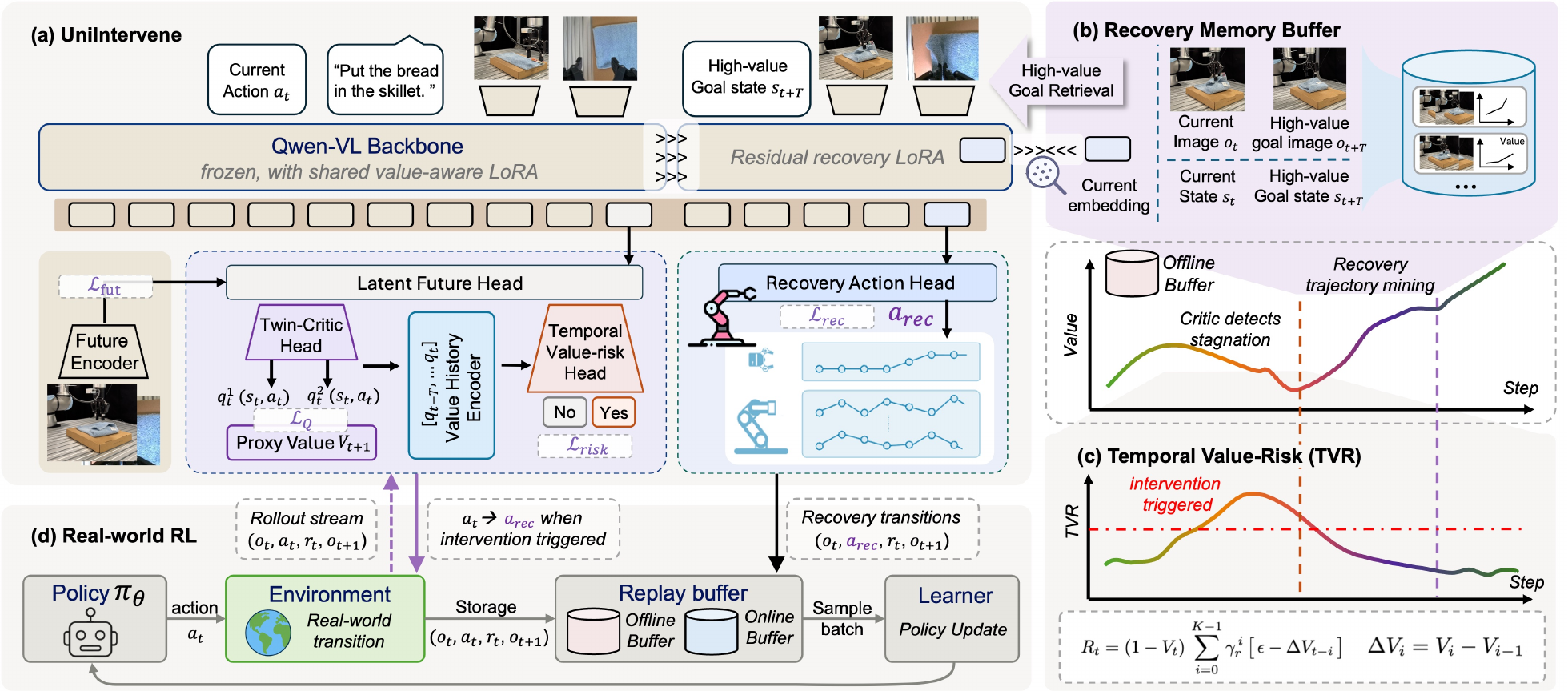}
  \caption{\textbf{Overview of \method{}.}
\textbf{(a)} A Qwen-VL backbone feeds a Latent Future Head with twin-critic and temporal value-risk supervision, and a Recovery Action Head producing corrective actions $a_{\mathrm{rec}}$.
\textbf{(b)} A memory buffer pairs past intervention states with retrieved high-value goals.
\textbf{(c)} The temporal value-risk $R_t$ triggers intervention upon sustained stagnation.
\textbf{(d)} \method{} plugs into the real-world RL loop, overriding $a_t$ with $a_{\mathrm{rec}}$ when triggered and contributing recovery transitions to the replay buffer.}
  \label{fig:method}
\end{figure}

We first formulate the autonomous intervention problem in real-world HiL-RL (Sec.~\ref{sec:problem}), then introduce future-conditioned action-value estimation as the basis for evaluating ongoing actions (Sec.~\ref{sec:future_value}). On top of this estimator, a temporal value-risk critic determines when intervention should be triggered (Sec.~\ref{sec:tvr}), and a memory-guided goal-conditioned recovery policy generates corrective actions to return the rollout to a productive state (Sec.~\ref{sec:recovery}). The overall pipeline is shown in Fig.~\ref{fig:method}.

\subsection{Problem Statement}
\label{sec:problem}
We consider real-world human-in-the-loop reinforcement learning for language-conditioned manipulation.
At each step $t$, the robot observes $o_t \in \mathcal{O}$, receives a task instruction $\ell \in \mathcal{L}$, and executes an action $a_t \in \mathcal{A}$ from a policy $\pi_\theta(a_t \mid o_t, \ell)$.
A human operator may intervene when the robot is unlikely to make meaningful task progress.
The objective is to maximize task return while minimizing the number of human interventions during training:
\begin{equation}
    \max_{\pi_\theta}\ 
    \mathbb{E}_{\tau \sim \pi_\theta}
    \!\left[
        \sum_{t=0}^{T}
        \gamma^{t}\, r(o_t, a_t, \ell)
        \;-\;
        \lambda_{\mathrm{int}}\, y_t
    \right],
    \label{eq:hil_objective}
\end{equation}
where $T$ is the episode horizon, $y_t \in \{0,1\}$ indicates whether a human intervention occurs at step $t$, $r(\cdot)$ is the task reward, $\gamma$ is the discount factor, and $\lambda_{\mathrm{int}}$ penalizes intervention cost.

In existing HiL-RL pipelines, $y_t$ is determined exogenously by the human operator: the human decides when to intervene and supplies the corrective actions, leaving the policy itself with no mechanism to recognize unproductive rollouts. We instead make this decision endogenous to the learning system. Concretely, we introduce a learned intervention module
\begin{equation}
    \mathcal{I}_\psi:
(o_t, \ell, a_t)
\mapsto
\big(\,
    \hat{q}_t,\;
    s_t^{\mathrm{int}},\;
    A_t^{\mathrm{rec}}\,
\big),
    \label{eq:intervention_mapping}
\end{equation}
which jointly predicts the value $\hat{q}_t$ of continuing the current action, an intervention score $s_t^{\mathrm{int}} \in [0,1]$, and a corrective action chunk $A_t^{\mathrm{rec}} = \{a_t^{\mathrm{rec}}, \ldots, a_{t+H}^{\mathrm{rec}}\}$. At deployment, the executed action follows $\pi_\theta(o_t, \ell)$ when $s_t^{\mathrm{int}} < \tau_{\mathrm{int}}$ and $a_t^{\mathrm{rec}}$ otherwise, with $\tau_{\mathrm{int}}$ a fixed threshold. This formulation turns intervention into an internal decision of the policy stack, with $\mathcal{I}_\psi$ jointly resolving when to interrupt the current action and how to recover.


\subsection{Future-conditioned Action-Value Estimation}
\label{sec:future_value}
The first component of $\mathcal{I}_\psi$ asks a single question: \emph{if the policy continues the current action, will the rollout move toward task completion?}
A model-free critic $Q(o_t, \ell, a_t)$ answers this only weakly in real-world manipulation, where rewards are sparse and a single frame reveals little about the action's downstream effect.
We instead take a model-based view, predicting the action-conditioned latent consequence and evaluating it with a value head trained against a robust offline target.

\paragraph{Latent consequence prediction.}
A vision-language backbone encodes the current observation, instruction, and action into a joint representation $h_t = f_{\mathrm{vlm}}(o_t, \ell, a_t)$, from which a future head predicts the latent consequence one step ahead, supervised against a frozen visual encoder $E_{\mathrm{vis}}$:
\begin{equation}
    \hat{z}_{t+1} = f_{\mathrm{fut}}(h_t), 
    \qquad
    \mathcal{L}_{\mathrm{fut}}
    = d\!\left(\hat{z}_{t+1},\, E_{\mathrm{vis}}(o_{t+1})\right),
    \label{eq:future_loss}
\end{equation}
where $d(\cdot,\cdot)$ is a latent distance, with $E_{\mathrm{vis}}$ instantiated as V-JEPA2~\cite{assran2025vjepa2}. This yields a forward-looking representation that aggregates progress-relevant cues over the action horizon, which is particularly informative under sparse rewards where instantaneous observations carry little task signal.

\paragraph{Value head and proxy supervision.}
A twin value head $f_{\mathrm{Q}}$ maps $\hat{z}_{t+1}$ to $\hat{q}_t = \min(\hat{q}_{1,t}, \hat{q}_{2,t})$ and is trained with a Smooth-$L_1$ loss against a target $q_t^\star$:
\begin{equation}
    \mathcal{L}_{\mathrm{Q}}
    = \sum_{k=1}^{2}
    \rho_{\beta}\!\left(\hat{q}_{k,t} - q_t^\star\right).
    \label{eq:q_loss}
\end{equation}
Since the value head drives both the trigger (Sec.~\ref{sec:tvr}) and the recovery-target retrieval (Sec.~\ref{sec:recovery}), we obtain $q_t^\star$ from a \emph{proxy value function} pre-trained offline on successful and failed trajectories with
$\mathcal{L}_{\mathrm{proxy}} = \mathcal{L}_{\mathrm{TD}} + \mathcal{L}_{\mathrm{prog}} + 0.05\,\mathcal{L}_{\mathrm{CQL}}$,
combining Bellman consistency, monotonic progress regression, and a small CQL term~\cite{kumar2020conservative} to suppress out-of-distribution overestimation, yielding a progress-aligned signal decoupled from the online loop (see Appendix for details).
\subsection{Temporal Value-Risk Intervention Trigger}
\label{sec:tvr}
The second component of $\mathcal{I}_\psi$ maps the estimated value into an intervention score $s_t^{\mathrm{int}}$. A naive rule that triggers whenever $\hat{q}_t$ is low is unreliable, because real-world manipulation frequently traverses transient low-value states during contact, alignment, or regrasping that do not warrant intervention. The decision should therefore depend not on the instantaneous value but on its \emph{temporal trend}: an intervention is warranted only when the policy is failing to accumulate value over time, a pattern that a single-step estimate cannot capture but that recent value dynamics reveal.

\paragraph{Temporal value-risk.}
We formalize this trend as the cumulative shortfall of recent value progress relative to an expected progress rate $\epsilon$. Let $\Delta V_i = V_i - V_{i-1}$ denote the one-step value change; a healthy rollout maintains $\Delta V_i \geq \epsilon$, whereas a stagnating one accumulates a positive shortfall $\epsilon - \Delta V_i$. We aggregate this over a window of length $K$ and modulate by the distance to success:
\begin{equation}
    R_t =
    \underbrace{(1 - V_t)}_{\text{progress remaining}}\,
    \sum_{i=0}^{K-1}
    \gamma_r^{\,i}\,
    \underbrace{\big[\,\epsilon - \Delta V_{t-i}\,\big]}_{\text{per-step shortfall}},
    \label{eq:tvr}
\end{equation}
where $\gamma_r \in (0,1)$ discounts older transitions so that $R_t$ reflects the recent trend. This yields a principled risk signal: $R_t$ grows large only when value increments persistently fall below $\epsilon$ across the window, while $(1 - V_t)$ attenuates the risk near task completion, so $R_t$ responds to sustained stagnation rather than isolated value drops.

\paragraph{Risk critic.}
To enable online prediction without recomputing the window at every step, a risk head $f_{\mathrm{risk}}$ regresses $R_t$ from the predicted future latent and an encoding of the recent value sequence $\mathcal{V}_t = [V_{t-K},\ldots,V_t]$:
\begin{equation}
    \hat{R}_t = f_{\mathrm{risk}}\!\left(\hat{z}_{t+1},\, f_{\mathrm{hist}}(\mathcal{V}_t)\right),
    \qquad
    \mathcal{L}_{\mathrm{risk}} = \rho_{\beta}(\hat{R}_t - R_t).
    \label{eq:risk_loss}
\end{equation}
We obtain the intervention score as $s_t^{\mathrm{int}} = \sigma(\hat{R}_t)$ and trigger recovery when $s_t^{\mathrm{int}} \ge \tau_{\mathrm{int}}$, so that the decision is grounded in the predicted trend of task progress rather than the instantaneous value.

\subsection{Memory-guided Goal-conditioned Recovery}
\label{sec:recovery}
The third component of $\mathcal{I}_\psi$ produces the action chunk $A_t^{\mathrm{rec}}$ when intervention is triggered. Generating corrective actions from a low-value state is under-specified: the critic indicates \emph{that} the rollout is unproductive, but not \emph{which} productive state to return to. We resolve ambiguity by decomposing recovery into target selection and target-conditioned control, grounding the former in experience.

\paragraph{Recovery memory.}
We maintain an offline memory $\mathcal{M} = \{(o_j^{\mathrm{fail}}, s_j^{\mathrm{fail}}, \ell_j, o_j^{g}, s_j^{g})\}_j$ aggregated from prior rollouts, where each entry pairs an intervention state (observation $o_j^{\mathrm{fail}}$ and state $s_j^{\mathrm{fail}}$) with a high-value future state ($o_j^{g}$, $s_j^{g}$) reached later in the same rollout, under instruction $\ell_j$. We admit an entry only when the proxy value of the future state exceeds a threshold, so that retrieval targets are guaranteed to be high-value under the same criterion that governs the trigger. This makes $\mathcal{M}$ a repository of verified recovery targets rather than arbitrary past states. Further details on memory construction and threshold selection are provided in Appendix.

\paragraph{Target retrieval and goal-conditioned recovery.}
Recovery from semantically similar failures should aim at similar productive states; we therefore retrieve a target by matching the current context to memory keys in a shared embedding space $\phi(\cdot)$:
\begin{equation}
    j^\star = \arg\max_j\; \mathrm{sim}\!\left(\phi(o_t, \ell),\, \phi(o_j^{\mathrm{fail}}, \ell_j)\right), \qquad g_t = (o_{j^\star}^{g},\, s_{j^\star}^{g}),
    \label{eq:memory_retrieval}
\end{equation}
where $\mathrm{sim}(\cdot,\cdot)$ denotes cosine similarity in the normalized embedding space, and $g_t = (o_t^{g}, s_t^{g})$ denotes the retrieved goal observation and state. Since $g_t$ rarely coincides with the current state, the past trajectory toward $g_t$ cannot be replayed directly; we instead train a goal-conditioned policy $\pi_{\mathrm{rec}}$ that generalizes to the current state, supervised by behavior cloning on rollout segments connecting each $o_j^{\mathrm{fail}}$ to its goal state. The recovery actions are decoded with a FAST tokenizer~\cite{pertsch2025fastefficientactiontokenization}, representing the corrective chunk as discrete frequency-domain tokens trained by per-token classification:
\begin{equation}
    \mathcal{L}_{\mathrm{rec}} = -\sum_{i=0}^{H} \log \pi_{\mathrm{rec}}\!\left(a_{t+i}^{\mathrm{rec}} \,\big|\, o_t, g_t, \ell,\, a_{t:t+i-1}^{\mathrm{rec}}\right).
    \label{eq:recovery_loss}
\end{equation}
Conditioning on $g_t$ converts the ill-posed recovery into a goal-reaching task: memory supplies \emph{where} to recover, while the policy learns \emph{how} to reach it.
\section{Experimental Results}
\label{sec:experiments}

We evaluate \method on robotic manipulation tasks to answer three key questions:
(1) Can \method improve task success while reducing human interventions compared with SOTA HiL-RL baselines (Sec.~\ref{subsec:comp_with_sota})?
(2) How much does each key component contribute to performance (Sec.~\ref{sec:ablation})?
(3) Does \method learn intervention and recovery behaviors, including detecting sustained low-value stagnation and retrieving recovery targets (Sec.~\ref{sec:analysis})?

\subsection{Experiment Setup}
\label{sec:exp_setup}

\noindent\textbf{Real-world Benchmark.}
We evaluate \method on five real-world manipulation tasks with a UR7e robotic arm, where human interventions are provided via a SpaceMouse. The benchmark covers multi-object interaction, contact-rich manipulation, and non-rigid object manipulation, including \textit{Pick Eggplant}, \textit{Tube Insertion}, \textit{RAM Insertion}, \textit{Wipe Whiteboard}, and \textit{Fold Towel}. These tasks require robust handling of pose variation, precise contact, long-horizon correction, and recovery from low-value states. Detailed settings are provided in Appendix.

\noindent\textbf{Baselines.}
We compare \method with three baselines for human-in-the-loop manipulation.
First, we use imitation learning (IL) as the policy baseline, where a $\pi_{0.5}$~\cite{black2025pi_} policy is fine-tuned with supervised fine-tuning (SFT) using 20 demonstrations per task.
Second, we adopt HIL-SERL~\cite{luo2024precise}, which trains manipulation policies with corrective interventions and off-policy RL updates.
Third, we compare with Failure-aware RL~\cite{li2026failure}, which combines failure prediction and recovery learning to reduce unsafe or unproductive exploration.
All baselines and \method are evaluated under the same robot platform, task initialization protocol, and SpaceMouse-based intervention interface. Detailed baseline settings are provided in Appendix.

\noindent\textbf{Evaluation Metrics.}
We report two main metrics: final success rate (SR) and human intervention rate (IR).
SR is measured over 20 evaluation episodes per task after training and reported as the percentage of successful episodes.
IR is computed over three independent training runs as the ratio between the total number of intervention steps and the total number of robot interaction steps.

\subsection{Comparisons with the State-of-the-Art}
\label{subsec:comp_with_sota}

\begin{table*}[t]
\newcolumntype{Y}{>{\centering\arraybackslash}X}
\centering
\small
\setlength{\tabcolsep}{3pt}
\caption{\textbf{Real-world Manipulation Results.}
\method achieves the best success rate on every task and the lowest intervention rate overall, improving average success by 8.6\% while cutting human interventions by 57\% relative to HiL-SERL.
SR: success rate (\%); IR: human intervention rate (\%). Best results in \textbf{bold}.}
\label{tab:real_world_results}
\renewcommand{\arraystretch}{1.2}
\begin{tabularx}{\textwidth}{l *{12}{Y}}
\toprule
\multirow{3}{*}{\textbf{Method}}
& \multicolumn{2}{c}{\textbf{Multi-object}}
& \multicolumn{6}{c}{\textbf{Contact-rich}}
& \multicolumn{2}{c}{\textbf{Non-rigid}}
& \multirow{3}{*}{\makecell{\textbf{Average}}}  \\
\cmidrule(lr){2-3} \cmidrule(lr){4-9} \cmidrule(lr){10-11}
& \multicolumn{2}{c}{\makecell{Pick\\Eggplant}}
& \multicolumn{2}{c}{\makecell{Tube\\Insertion}}
& \multicolumn{2}{c}{\makecell{RAM\\Insertion}}
& \multicolumn{2}{c}{\makecell{Wipe\\Whiteboard}}
& \multicolumn{2}{c}{\makecell{Fold\\Towel}}
&  \\
\cmidrule(lr){2-3} \cmidrule(lr){4-5} \cmidrule(lr){6-7}
\cmidrule(lr){8-9} \cmidrule(lr){10-11}
& \multicolumn{1}{c}{SR\,\textcolor{teal}{$\uparrow$}} & \multicolumn{1}{c}{IR\,\textcolor{purple}{$\downarrow$}}
& \multicolumn{1}{c}{SR\,\textcolor{teal}{$\uparrow$}} & \multicolumn{1}{c}{IR\,\textcolor{purple}{$\downarrow$}}
& \multicolumn{1}{c}{SR\,\textcolor{teal}{$\uparrow$}} & \multicolumn{1}{c}{IR\,\textcolor{purple}{$\downarrow$}}
& \multicolumn{1}{c}{SR\,\textcolor{teal}{$\uparrow$}} & \multicolumn{1}{c}{IR\,\textcolor{purple}{$\downarrow$}}
& \multicolumn{1}{c}{SR\,\textcolor{teal}{$\uparrow$}} & \multicolumn{1}{c}{IR\,\textcolor{purple}{$\downarrow$}}
& \multicolumn{1}{c}{SR\,\textcolor{teal}{$\uparrow$}} & \multicolumn{1}{c}{IR\,\textcolor{purple}{$\downarrow$}} \\
\midrule
$\pi_{0.5}$ (SFT)
& \textbf{95} & --   & 30 & --   & 10 & --   & 65 & --   & 70 & --   & 54 & -- \\
HiL-SERL
& 90 & 28.7 & 60 & 30.2 & 85 & 32.3 & 85 & 30.5 & 85 & 49.8 & 81 & 34.3 \\
HiL-SERL + FA-RL
& 85 & 20.4 & 60 & 22.1 & 75 & 27.9 & 80 & 21.9 & 85 & 30.9 & 77 & 24.6 \\
\midrule
HiL-SERL + \method
& \textbf{95} & \textbf{10.0} & \textbf{70} & \textbf{15.8} & \textbf{95} & \textbf{12.1}
& \textbf{90} & \textbf{10.9} & \textbf{90} & \textbf{24.1} & \textbf{88} & \textbf{14.6} \\
\bottomrule
\vspace{-0.8cm}
\end{tabularx}
\end{table*}

As shown in Table~\ref{tab:real_world_results}, \method{} achieves a comprehensive win, attaining the best success rate (SR) on every task (88\% on average) while requiring the fewest human interventions (IR of 14.6\%, less than half that of HiL-SERL), improving task performance and training efficiency simultaneously. The offline $\pi_{0.5}$ (SFT) policy performs reasonably on simple pick-and-place but degrades sharply on contact-rich insertion (30\% and 10\% on Tube and RAM Insertion), confirming that demonstrations alone do not cover the fine-grained corrections these tasks require. HiL-SERL recovers this gap through online intervention, but at a high cost: its average IR reaches 34.3\%, peaking at 49.8\% on Fold Towel where deformable dynamics demand frequent corrections.

FA-RL reduces intervention by predicting failures, yet its trigger is tied to explicit failure signals and is therefore unreliable when failures are visually subtle. This is most evident on RAM Insertion and Wipe Whiteboard, where small inter-frame appearance changes make failure hard to detect, leaving SR at 75\% and 80\%, below HiL-SERL's 85\%, while still consuming substantial intervention. \method{} instead triggers on the temporal trend of estimated value rather than on discrete failure events, detecting stagnation more sensitively and redirecting the rollout toward a retrieved high-value target. As a result, \method{} attains the best SR on every task while lowering IR to 14.6\%, less than half of HiL-SERL's 34.3\% and below FA-RL's 24.6\%. The gains are largest on the contact-rich tasks where prior triggers struggle, indicating that value-trend monitoring provides a more robust intervention signal than failure prediction.


\subsection{Analysis of Learned Intervention and Recovery Behaviors}
\label{sec:analysis}

\begin{figure}[t]
  \centering
  \includegraphics[width=1.0\linewidth]{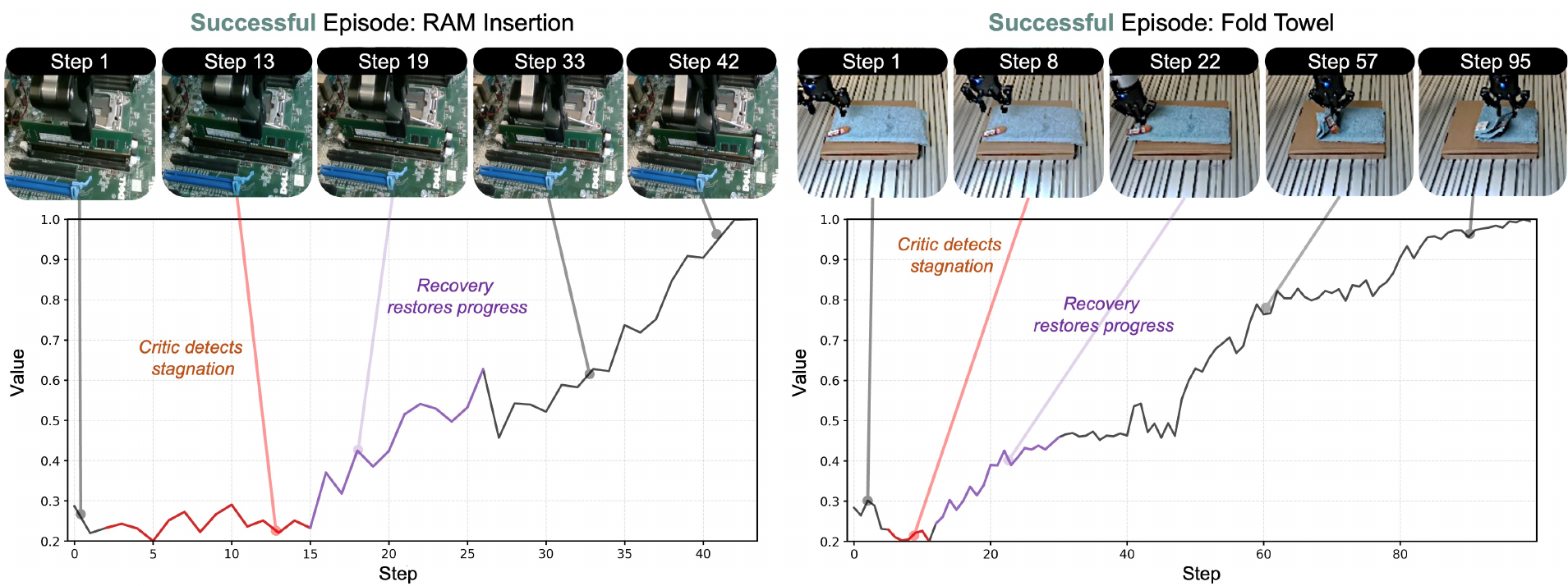}
\caption{\textbf{\method enables value-aware recovery in real-world tasks.}
Case studies on RAM Insertion and Fold Towel show that \method detects sustained low-value stagnation, triggers corrective recovery, and restores task progress toward successful completion.}
\vspace{-0.5cm}
\label{fig:case_study}
\end{figure}

To examine whether \method{} learns meaningful behavior beyond aggregate metrics, we visualize the predicted value over two successful episodes in Fig.~\ref{fig:case_study}. Both follow a consistent stagnation-trigger-recovery pattern. Early on, the predicted value stays low and flat, reflecting genuine stagnation rather than frame-level noise; \method{} waits until the stagnation is sustained before triggering, confirming that the trigger responds to a low-value \emph{trend} rather than instantaneous fluctuations. Notably, the trigger fires at the local value minimum (step 13 on RAM Insertion, step 8 on Fold Towel) rather than at the first low-value frame, indicating that the temporal aggregation in $R_t$ correctly distinguishes the bottom of stagnation from incidental dips. After triggering, the goal-conditioned policy steers the rollout toward a retrieved high-value target, and the value rises smoothly and monotonically to completion without secondary triggering, indicating that a single retrieved goal is sufficient to restore productive progress and that the targets provide feasible intermediate goals. The two tasks differ in horizon (42 vs.\ 95 steps) yet share the same structure, suggesting the learned behavior generalizes across manipulation types.

\subsection{Ablation Studies}
\label{sec:ablation}

\begin{wraptable}{r}{0.56\textwidth}
\vspace{-0.8em}
\begin{minipage}{\linewidth}
\centering
\footnotesize
\setlength{\tabcolsep}{4pt}
\renewcommand{\arraystretch}{1.05}
\caption{\textbf{Ablation Study.} We ablate the key components of \method{} on \textit{Pick Eggplant} and \textit{RAM Insertion}.}
\label{tab:ablation}
\begin{tabular}{lcccc}
\toprule
\multirow{2}{*}{\textbf{Variant}}
& \multicolumn{2}{c}{\textbf{Value-Risk Critic}}
& \multicolumn{2}{c}{\textbf{Online RL}} \\
\cmidrule(lr){2-3}
\cmidrule(lr){4-5}
& \makecell{$Q$-Loss\\$\downarrow$}
& \makecell{Int. F1\\$\uparrow$}
& SR\,(\%)\textcolor{green}{$\uparrow$}
& IR\,(\%)\textcolor{purple}{$\downarrow$} \\
\midrule

w/o Future Pred.      & 0.005 & 0.878 & 90 & 15.8 \\
w/o Value Pred.       & -- & 0.845 & 85 & 18.7 \\
w/o TVR               & 0.004 & 0.832 & 85 & 16.9 \\
w/o Memory Goal  & 0.004 & 0.882 & 85 & 16.1 \\

\midrule
\rowcolor{\ourcolor}
\textbf{\method}      & \textbf{0.004} & \textbf{0.882} & \textbf{95} & \textbf{11.1} \\

\bottomrule
\end{tabular}
\end{minipage}
\vspace{-1em}
\end{wraptable}
Table~\ref{tab:ablation} ablates the key components of \method{} on two representative tasks, \textit{Pick Eggplant} and \textit{RAM Insertion}, using both offline metrics ($Q$-loss, intervention F1) and online performance (SR, IR).
The Value-Risk Critic is the most critical module: removing value prediction yields the lowest F1 (0.845) and highest IR (18.7\%); the absent $Q$-loss entry in the table reflects that no value head is trained, confirming that reliable value estimation is essential for intervention triggering.
Removing the temporal value-risk (TVR) trigger causes a comparable drop (F1 $0.832$, IR $16.9\%$), as the model no longer aggregates value dynamics over time and instead reacts to transient value drops.
In contrast, removing future prediction or the memory goal keeps offline F1 nearly intact ($0.878$ and $0.882$), but still lowers online SR from $95\%$ to $90\%$ and $85\%$, indicating that these components mainly affect recovery quality rather than trigger classification.
This also suggests that offline intervention F1 alone is insufficient to characterize recovery effectiveness, since memory-guided goal selection can substantially affect online success without changing trigger accuracy.
Overall, on these two ablation tasks, the full model achieves the best results across all metrics ($95\%$ SR, $11.1\%$ IR), showing that value-risk estimation drives intervention decisions while future prediction and memory-guided recovery improve online recovery.


\section{Conclusion}
\label{sec:conclusion}
We presented \method{}, an agentic intervention model that internalizes the intervention decision in real-world human-in-the-loop reinforcement learning. Rather than relying on human operators to recognize unproductive rollouts after the fact, \method{} couples future-conditioned action-value estimation with temporal value-risk modeling to detect stagnation as it emerges, and resolves the recovery ambiguity by retrieving high-value targets from past intervention episodes and producing goal-conditioned corrective actions. Across diverse real-world manipulation tasks, this design improves task success while substantially reducing the human interventions consumed during online training. More broadly, our results suggest that the dominant cost in real-world HiL-RL is not catastrophic failure but slow, unproductive exploration, and that this cost can be absorbed by a learned value-aware recovery process, pointing toward a path in which human supervision is reserved for the residual cases that learned recovery cannot resolve.

\section{Limitations}

Despite its effectiveness, \method{} has several limitations.
First, the intervention trigger depends on the quality of the proxy value function; if the value estimate is poorly calibrated or fails to reflect task progress, the temporal value-risk critic may trigger recovery too early or too late.
Second, the goal-conditioned recovery policy relies on a memory of past intervention episodes, so retrieval-based recovery may be less reliable for failure modes not covered by the memory.
Third, our experiments focus on tabletop manipulation with a single robot embodiment; scaling \method{} to diverse robots, long-horizon mobile manipulation, and multi-task deployment will require richer recovery memories and more robust value estimation.
Finally, \method{} reduces but does not eliminate the need for human supervision, which remains important for rare, unsafe, or out-of-distribution failures that the learned recovery policy cannot resolve.

\bibliographystyle{plain}
\bibliography{references}
\clearpage
\appendix
\section*{Appendix}
%
%

\providecommand{\method}{\textsc{UniIntervene}}

\appendix

\section{Overview}
\label{appendix:overview}

This supplement details the full method behind \method, from the proxy value
function used to score rollouts, through the temporal value-risk trigger, to the
memory-guided recovery policy. We first describe the real-world benchmark and hardware
(Appendix~\ref{appendix:tasks}), then give the implementation of each component in the
order they appear in the main paper: the proxy value function and its labeling
(Appendix~\ref{appendix:proxyvf}), the future-conditioned action-value estimator and its
network architecture (Appendix~\ref{appendix:vq1}), the temporal value-risk trigger and
its offline mining rules (Appendix~\ref{appendix:tvr}), and the recovery memory together
with the goal-conditioned recovery policy (Appendix~\ref{appendix:recovery}). We then
list the end-to-end training procedure and all hyperparameters
(Appendix~\ref{appendix:training}), the baseline configurations
(Appendix~\ref{appendix:baselines}), and qualitative results on the learned behavior
(Appendix~\ref{appendix:results}).

A short video accompanies this submission. It walks through a single real-world rollout
in which \method\ detects sustained value stagnation during a contact-rich insertion, takes
over from the policy, retrieves a high-value recovery target, and returns the rollout to a
productive state without human intervention.

\section{Real-World Benchmark and Hardware}
\label{appendix:tasks}

\subsection{Hardware Setup}
\begin{wrapfigure}{r}{0.50\linewidth}
  \centering
  \vspace{-8pt}
  \includegraphics[width=\linewidth]{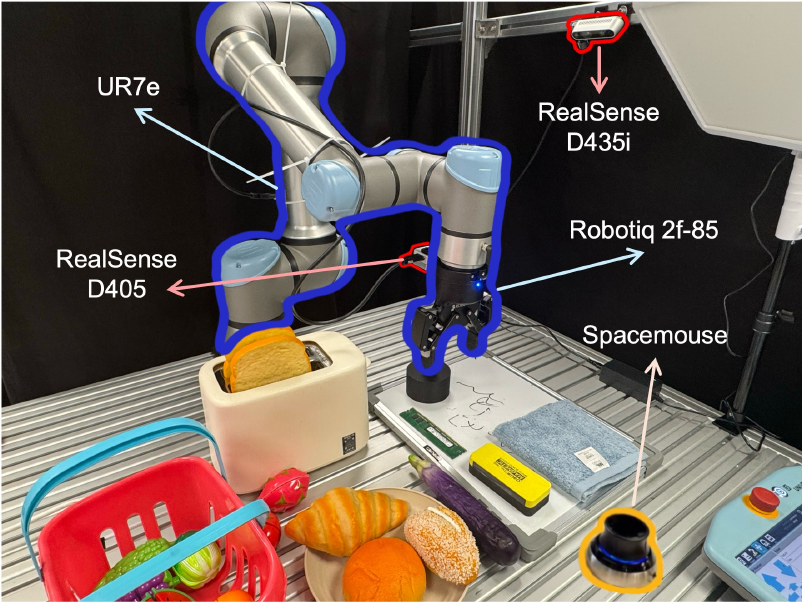}
  \caption{Real-world hardware setup. A UR7e arm with a parallel-jaw gripper, a wrist camera and a fixed third-person camera, and a SpaceMouse for corrective takeover.}
  \label{fig:hardware}
\end{wrapfigure}
All experiments run on a single UR7e robotic arm with a parallel-jaw gripper. Visual
observations come from two calibrated RGB cameras, one mounted on the wrist and one fixed
to provide a third-person workspace view, both resized to $224\times224$ before being fed
to the policy and the intervention module. Human interventions are supplied through a
3D-connexion SpaceMouse, which lets the operator override the policy with a corrective
end-effector velocity command whenever a takeover is needed. The same platform, camera
placement, and intervention interface are shared by \method\ and all baselines, so that
differences in measured intervention rate reflect the method rather than the teleoperation
setup. Figure~\ref{fig:hardware} shows the platform.

\subsection{Task Suite}
\label{appendix:task_list}
\begin{figure}[h]
  \centering
  \includegraphics[width=\linewidth]{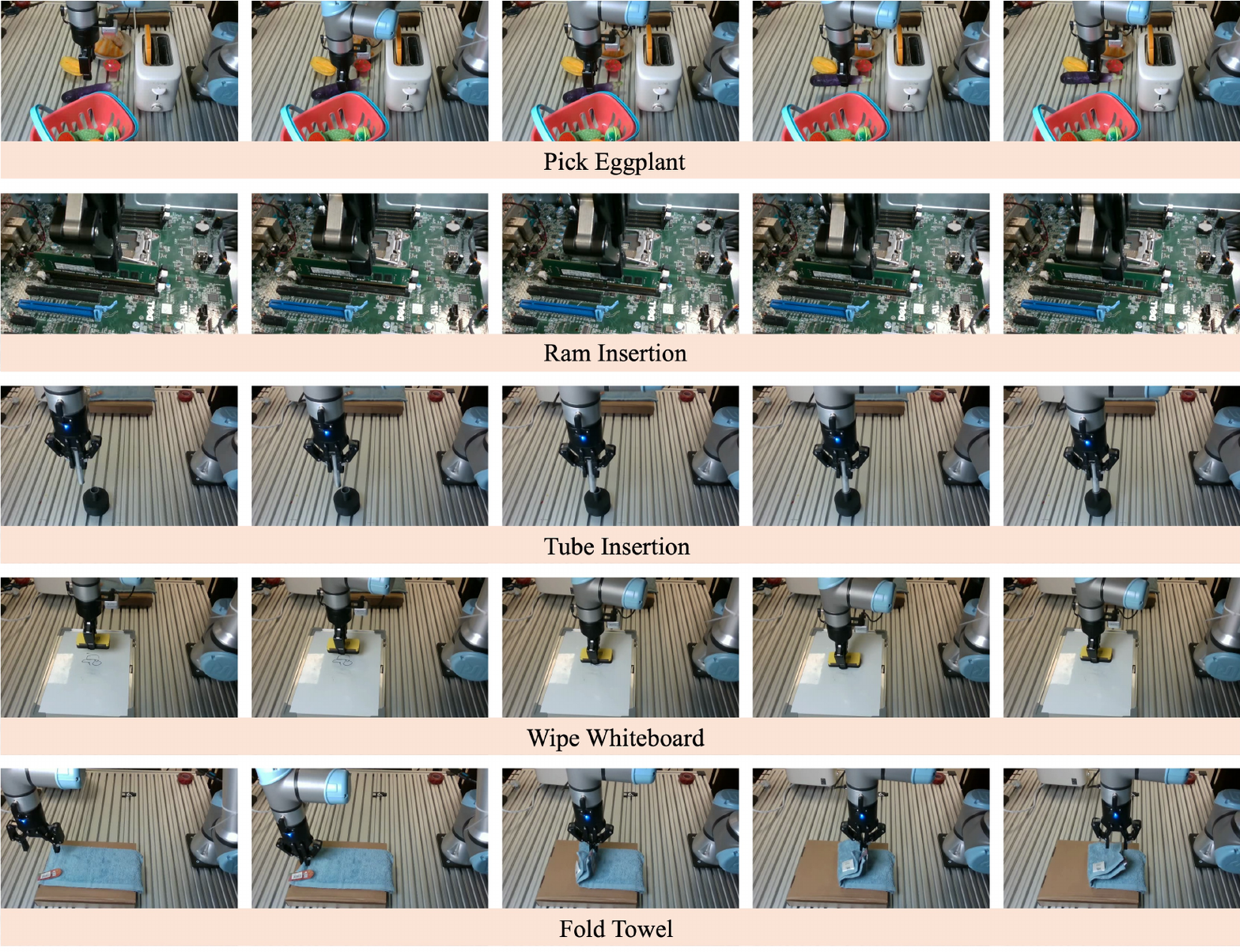}
  \caption{The five real-world manipulation tasks used to evaluate \method, shown from the
  third-person workspace camera during execution.}
  \label{fig:tasks}
\end{figure}

The benchmark contains five tasks chosen to span multi-object interaction, contact-rich
assembly, and non-rigid object manipulation. Each task is initialized with randomized
object poses, and an episode is scored as a success only if the final configuration
satisfies the task-specific criterion below. Figure~\ref{fig:tasks} shows a snapshot of
the five tasks.

\textbf{Pick Eggplant (multi-object).} An eggplant and three or four distractor objects are
placed at randomized positions. The arm must grasp the eggplant and lift it clear of the
workspace. Success requires a stable grasp on the correct object with the distractors left
undisturbed.

\textbf{Tube Insertion (contact-rich precision manipulation).} A flexible tube must be aligned with and inserted
into a fixed port. The task demands sub-centimeter alignment and tolerates only small
lateral error before the tube buckles. Success requires the tube tip to be seated inside
the port.

\textbf{RAM Insertion (contact-rich precision manipulation).} A memory module must be aligned to a slot and pressed
home. Small appearance changes between a near-miss and a seated module make failure visually
subtle, which is the regime where failure-prediction triggers tend to break down. Success
requires the module to be fully seated and latched.

\textbf{Wipe Whiteboard (contact-rich).} The arm holds an eraser
and must wipe a marked region on a vertical whiteboard while maintaining contact force.
Success requires the marked region to be cleared without losing surface contact.

\textbf{Fold Towel (non-rigid).} A towel laid flat must be folded along a target line. The
deformable dynamics produce frequent low-value states during regrasping and re-alignment,
and demonstrations rarely cover the corrections needed. Success requires the towel to be
folded into the target shape with aligned edges.

\section{Proxy Value Function}
\label{appendix:proxyvf}

The proxy value function provides the offline target $q^\star_t$ that supervises the value
head in the main module (Eq.~4 of the main paper). Because both the trigger and the
recovery-target retrieval are driven by this signal, it is trained separately from the
online loop and frozen during \method\ training.

\subsection{Progress Labeling}
\label{appendix:nttg}
We assign a normalized time-to-go (NTTG) progress label to every transition before
training the value function. For a successful episode of length $T$, transition $t$
receives a progress label $v_t = t/(T{-}1)$, so that value increases monotonically from
$0$ at the start to $1$ at task completion. Failed episodes receive $v_t = 0$ throughout,
together with a large negative raw return that suppresses their value estimate. This
labeling turns sparse terminal success into a dense per-step progress signal, and it is the
target that the progress regression term below regresses against.

\subsection{Architecture}
\label{appendix:vf_arch}
The proxy value function is a compact stitched vision-language model. A frozen
SigLIP-SO400M vision tower encodes the observation into a $1152$-dimensional feature, a
two-layer MLP projector maps it to the $640$-dimensional language space, and a
Gemma-3-270M language model fuses it with the instruction. Twin value heads read the
fused representation and predict a scalar value, and the reported value is the minimum of
the two heads, $V(o,\ell)=\min(V_1,V_2)$, which guards against single-head overestimation.
The model is trained in two stages. The first stage aligns the projector on
general vision-language data with the vision tower and language model frozen. The second
stage unfreezes the projector and value heads on robot data with the vision tower kept
frozen.

\subsection{Training Objective}
\label{appendix:vf_loss}
We pre-train the proxy value function offline on a dataset $\mathcal{D}$ of recorded
transitions $(o_t,\ell,r_t,o_{t+1},d_t)$, drawn from successful trajectories $\mathcal{D}^{+}$
and failed trajectories $\mathcal{D}^{-}$, where $r_t$ is the sparse terminal reward ($1$ at a
successful terminal state and $0$ otherwise) and $d_t$ is the done flag. The network carries
twin value heads and we read the value as $V_\phi(o,\ell)=\min_{k\in\{1,2\}}V_{\phi,k}(o,\ell)$.
This $V_\phi$ is the quantity the main paper writes as $V_t \equiv V_\phi(o_t,\ell)$ in the
temporal value-risk of Sec.~3.3, and it supplies the regression target
$q^\star_t \equiv V_\phi(o_{t+1},\ell)$ for the value head in Eq.~4. The proxy objective is
\begin{equation}
\mathcal{L}_{\text{proxy}} = \lambda_{\text{TD}}\,\mathcal{L}_{\text{TD}}
  + \underbrace{\lambda_{\text{label}}\,\mathcal{L}_{\text{label}}
  + \lambda_{\text{prog}}\,\mathcal{L}_{\text{mono}}}_{\textstyle \mathcal{L}_{\text{prog}}}
  + \lambda_{\text{CQL}}\,\mathcal{L}_{\text{CQL}},
\label{eq:vf_proxy}
\end{equation}
with weights $(\lambda_{\text{TD}},\lambda_{\text{label}},\lambda_{\text{prog}},\lambda_{\text{CQL}})=(1,\,0.3,\,1,\,0.05)$,
which recovers the grouped form $\mathcal{L}_{\text{TD}}+\mathcal{L}_{\text{prog}}+0.05\,\mathcal{L}_{\text{CQL}}$
stated in the main paper. Writing $\rho_\beta$ for the Smooth-L1 loss with $\beta=1$,
$\mathrm{sg}[\cdot]$ for the stop-gradient, and $v_t=t/(T{-}1)$ for the NTTG progress label of
Appendix~\ref{appendix:nttg}, the individual terms are
\begin{align}
\mathcal{L}_{\text{TD}} &= \sum_{k\in\{1,2\}}\mathbb{E}_{\mathcal{D}}\Big[\rho_\beta\big(V_{\phi,k}(o_t,\ell)
  - \mathrm{sg}\!\left[\,r_t + \gamma\,(1-d_t)\,V_\phi(o_{t+1},\ell)\,\right]\big)\Big], \label{eq:vf_td}\\
\mathcal{L}_{\text{label}} &= \mathbb{E}_{\mathcal{D}^{+}}\big[\rho_\beta\big(V_\phi(o_t,\ell)-v_t\big)\big], \label{eq:vf_label}\\
\mathcal{L}_{\text{mono}} &= \mathbb{E}_{\mathcal{D}^{+}}\big[\max\!\big(0,\;V_\phi(o_t,\ell)-V_\phi(o_{t+1},\ell)\big)\big], \label{eq:vf_mono}\\
\mathcal{L}_{\text{CQL}} &= \mathbb{E}_{o_t\sim\mathcal{D}}\Big[\log\!\!\sum_{\tilde o\in\mathcal{N}(o_t)}\!\!\exp V_\phi(\tilde o,\ell)\Big]
  - \mathbb{E}_{o_t\sim\mathcal{D}}\big[V_\phi(o_t,\ell)\big]. \label{eq:vf_cql}
\end{align}
Equation~\eqref{eq:vf_td} enforces one-step Bellman consistency on both heads against a
bootstrapped, stop-gradient target. Equations~\eqref{eq:vf_label} and~\eqref{eq:vf_mono} act
only on successful trajectories: the first regresses the value onto the NTTG progress label,
and the second is a monotonicity hinge that penalizes any drop in value between consecutive
successful steps, so that $V_\phi$ rises along productive rollouts. Equation~\eqref{eq:vf_cql}
is the state-value form of the Conservative Q-Learning regularizer. It lowers a soft-maximum
of the value over a set of negative or out-of-distribution candidate states $\mathcal{N}(o_t)$,
sampled from failed trajectories, while raising the value on the observed in-distribution
state, which caps the value assigned to states the data does not support and suppresses the
over-estimation that would otherwise mislead the trigger. We use a discount of $\gamma=0.99$,
a five-epoch progress warmup during which only Eqs.~\eqref{eq:vf_label} and~\eqref{eq:vf_mono}
are active, and balanced sampling of $\mathcal{D}^{+}$ and $\mathcal{D}^{-}$ so that the rarer
failure transitions are not drowned out. The per-term weights are repeated in
Table~\ref{tab:vf_config}.

\begin{table}[h]
\centering
\caption{Proxy value function configuration and training-objective weights.}
\label{tab:vf_config}
\begin{tabular}{lc@{\hskip 2.2em}lc}
\toprule
\textbf{Setting} & \textbf{Value} & \textbf{Setting} & \textbf{Value} \\
\midrule
Vision tower            & SigLIP-SO400M   & TD weight $\lambda_{\text{TD}}$      & $1.0$ \\
Language model          & Gemma-3-270M    & Progress label weight $\lambda_{\text{label}}$ & $0.3$ \\
Projector               & MLP $1152\!\to\!640$ & Progress weight $\lambda_{\text{prog}}$ & $1.0$ \\
Value heads             & twin, $\min(V_1,V_2)$ & Conservative weight $\lambda_{\text{CQL}}$ & $0.05$ \\
Discount $\gamma$       & $0.99$          & Progress warmup epochs               & $5$ \\
Success/fail sampling   & balanced ($0.5$) & Early-stop patience                 & $6$ \\
\bottomrule
\end{tabular}
\end{table}

\subsection{Value Function Quality}
\label{appendix:vf_quality}
\begin{figure}[h]
  \centering
  \includegraphics[width=\linewidth]{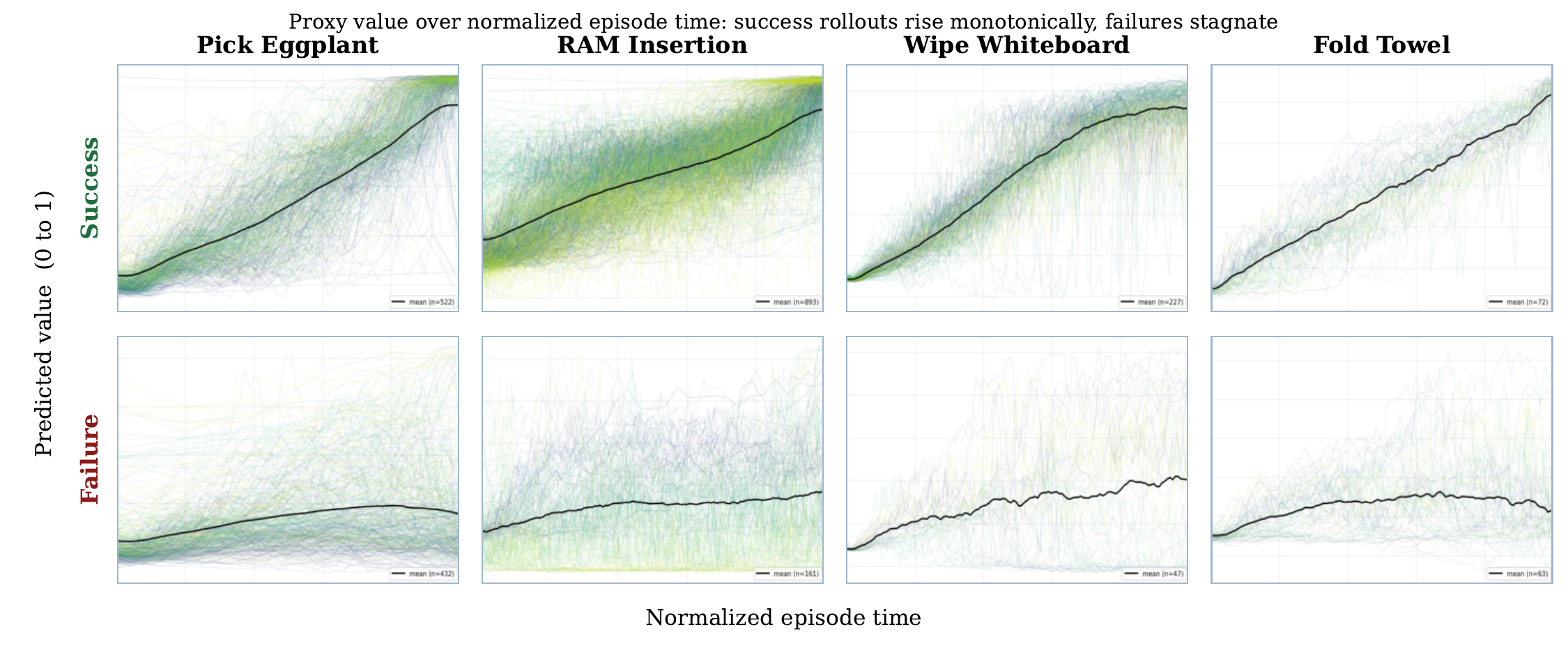}
  \caption{Predicted proxy value over normalized episode time, overlaid across all
  evaluation trajectories (faint lines) with the per-task mean in bold. Top row: successful
  episodes. Bottom row: failed episodes. Successful rollouts accumulate value monotonically
  while failed rollouts remain low and flat, and the gap is consistent across tasks of very
  different horizon.}
  \label{fig:vf_overlay}
\end{figure}

On held-out validation episodes the value heads cleanly separate successful from failed
rollouts. On the insertion split the mean predicted value reaches $0.53$ on successful
trajectories against $0.09$ on failed ones, and on Pick Eggplant the gap is $0.53$ against
$0.24$. The two heads agree closely, with a head disagreement below $3\times10^{-3}$ in all
splits, indicating that the conservative twin design does not collapse to a single mode.
Figure~\ref{fig:vf_overlay} overlays the predicted value against normalized episode time
for every successful and failed trajectory across four tasks. Successful rollouts rise
monotonically toward $1$, while failed rollouts stagnate at low value. This separation holds
across tasks of very different horizon, and it is what makes the value trend, rather than
the instantaneous frame, a usable progress signal for the trigger.

\section{Future-conditioned Action-Value Estimation}
\label{appendix:vq1}

This section details the network that realizes the intervention module $I_\psi$. A single
prompt-only forward pass of a vision-language backbone produces one shared hidden state from
which all heads read, so the future prediction, value estimation, and trigger are computed
without separate encoders.

\subsection{Backbone and Shared Hidden State}
\label{appendix:backbone}
The backbone is a Qwen3-VL-2B-Instruct model adapted with LoRA of rank $16$ and scaling
$\alpha=32$, applied to the attention projections \texttt{q\_proj}, \texttt{k\_proj},
\texttt{v\_proj}, and \texttt{o\_proj} with dropout $0.05$. We append a learned special
token \texttt{<VQ\_QUERY>} to the prompt and take its last-layer hidden state $h_t$ as the
shared representation. The future head, the value head, and the trigger head all consume
this same $h_t$. The value condition passed to the trigger head is detached before the
intervention head reads it, so the trigger loss cannot corrupt value calibration.

\subsection{Latent Future Prediction}
\label{appendix:future}
The future head is a two-layer MLP that expands $h_t$ to four times the hidden width and
projects it to the latent dimension, producing the predicted next-step latent
$\hat z_{t+1}$ in pooled form. The supervision target is a frozen V-JEPA2 encoder
(\texttt{vjepa2-vitl-fpc64-256}) applied to the realized next observation $o_{t+1}$, which
yields a $1024$-dimensional representation aggregated over $256$ tokens. We train the head
with a normalized MSE that L2-normalizes prediction and target before comparison, so that
the loss measures directional agreement in latent space rather than raw magnitude. This
forward-looking latent is the input to both the value head and the risk head, which is why
removing it degrades recovery quality even when trigger accuracy is preserved (Table~2 of
the main paper).

\subsection{Twin Value Head and Auxiliary Heads}
\label{appendix:heads}
The twin value head maps the predicted latent to two scalars and returns
$\hat q_t = \min(\hat q_{1,t}, \hat q_{2,t})$, trained with a Smooth-L1 loss against the
proxy target $q^\star_t$. An auxiliary current-value head regresses the proxy value of the
current step from $h_t$, which stabilizes the value scale early in training. The risk head
reads the predicted latent together with an encoding of the recent value sequence and
regresses the temporal value-risk target defined in Appendix~\ref{appendix:tvr}. The
intervention head reads the pooled latent concatenated with the detached value scalar and
outputs the trigger logit. Table~\ref{tab:vq1_arch} summarizes the architecture and the
training hyperparameters shared across tasks.

\begin{table}[h]
\centering
\caption{\method\ intervention module architecture and training settings.}
\label{tab:vq1_arch}
\resizebox{\textwidth}{!}{
\begin{tabular}{lc@{\hskip 2.2em}lc}
\toprule
\textbf{Component / Setting} & \textbf{Value} & \textbf{Component / Setting} & \textbf{Value} \\
\midrule
Backbone                  & Qwen3-VL-2B-Instruct       & Optimizer                 & AdamW \\
LoRA rank / $\alpha$ / dropout & $16$ / $32$ / $0.05$   & Backbone LR               & $2\times10^{-5}$ \\
LoRA targets              & q,k,v,o proj               & Head LR                   & $1\times10^{-4}$ \\
Shared token              & \texttt{<VQ\_QUERY>}       & Epochs                    & $30$ \\
Future target encoder     & V-JEPA2 ViT-L (frozen)     & Batch size                & $12$--$100$ \\
Future latent dim / tokens& $1024$ / $256$             & Future loss weight        & $1.0$ \\
Future latent mode        & pooled                     & Value (Q) loss weight     & $1.0$ \\
Value head                & twin, $\min(Q_1,Q_2)$      & Intervention loss weight  & $1.0$ \\
Value target              & proxy $q^\star_{t+1}$, Smooth-L1 & TVR loss weight     & $1.0$ \\
Value-history dim $K$     & $8$                        & Current-value loss weight & $1.0$ \\
History embed dim         & $32$                       & Precision                 & bf16 \\
\bottomrule
\end{tabular}
}
\end{table}

\section{Temporal Value-Risk Trigger}
\label{appendix:tvr}

The trigger maps the value estimate into an intervention score using the temporal trend of
value rather than its instantaneous level. We describe the offline labeling used to learn
the trend, the regression target, and the loss.

We label intervention points offline by scanning the per-episode proxy value sequence,
min-max normalized within each episode, with a sliding window of length $8$. A step is
marked as an intervention point if it satisfies any of three patterns. The \emph{decline}
rule fires when the windowed value shows a sustained negative slope with a sufficient
fraction of downward steps. The \emph{plateau} rule fires when the window is nearly flat,
measured by the fraction of steps whose absolute change stays below a small threshold and a
bound on the window amplitude. The \emph{abrupt-drop} rule marks the trough of a single
large drop that does not recover within a short horizon. 

Following Eq.~5 of the main paper, the temporal value-risk at step $t$ is the discounted
cumulative shortfall of recent value progress relative to an expected rate $\epsilon$,
attenuated by the remaining distance to success,
\[
R_t = (1 - V_t)\sum_{i=0}^{K-1}\gamma_r^{\,i}\bigl(\epsilon - \Delta V_{t-i}\bigr),
\qquad \Delta V_i = V_i - V_{i-1},
\]
with window $K=8$, expected progress rate $\epsilon=0.005$, and trend discount
$\gamma_r=0.9$. A value-history encoder embeds the recent value sequence and its one-step
differences into a $32$-dimensional vector, which the risk head consumes together with the
predicted future latent to regress $\hat R_t$. The intervention score is
$s^{\text{int}}_t = \sigma(\hat R_t)$, and recovery is triggered when
$s^{\text{int}}_t \ge \tau_{\text{int}}$ with $\tau_{\text{int}}=0.5$. The trigger head is
trained with a binary focal loss ($\alpha=0.75$, $\gamma=2.0$) against the mined labels,
which counteracts the heavy negative skew of intervention points without resorting to a
hand-tuned positive weight.

\section{Memory Construction and Goal-conditioned Recovery}
\label{appendix:recovery}

When the trigger fires, the recovery component selects a high-value target and produces a
corrective action chunk toward it. The target is grounded in an offline memory of verified
recovery segments.

\subsection{Recovery Memory Construction}
\label{appendix:bank}
We build the memory by mining value-improvement segments from prior rollouts. For each
episode we form candidate segments of a fixed span and keep those whose normalized value
rises by at least a delta threshold and whose start value is already reasonably high, so
that the segment endpoint is a genuinely high-value state under the same criterion that
governs the trigger. To avoid over-representing easy regions, candidates are bucketed into
progress bins and a fixed quota is sampled per bin, capped per episode. Each retained entry
pairs the segment start, treated as an intervention state, with its high-value endpoint,
treated as the recovery goal, under the task instruction. Table~\ref{tab:bank} reports the
resulting memory for each task. The candidate pools are large, between roughly three
thousand and forty thousand segments, from which a balanced set of $120$ to $240$ verified
targets is retained. Fold Towel uses a longer span and more progress bins because its
horizon is longer and its value trajectory is noisier.

\begin{table}[h]
\centering
\caption{Recovery memory statistics per task. Candidates are the mined value-improvement
segments; items are the retained, progress-balanced recovery targets. Bins$\times$quota is
the progress-bin sampling layout.}
\label{tab:bank}
\begin{tabular}{lccccccc}
\toprule
\textbf{Task} & \textbf{Candidates} & \textbf{Items} & \textbf{Bins$\times$Quota} & \textbf{Span} & \textbf{$\delta$ thr.} \\
\midrule
Pick Eggplant   & $11{,}879$ & $120$ & $10\times12$ & $8$  & $0.40$  \\
Tube Insertion  & $14{,}861$ & $120$ & $10\times12$ & $8$  & $0.40$  \\
RAM Insertion   & $40{,}742$ & $120$ & $10\times12$ & $8$  & $0.40$  \\
Wipe Whiteboard & $17{,}638$ & $120$ & $10\times12$ & $8$  & $0.40$  \\
Fold Towel      & $\phantom{0}3{,}574$  & $240$ & $20\times12$ & $16$ & $0.50$ \\
\bottomrule
\end{tabular}
\end{table}

\begin{figure}[h]
  \centering
  \includegraphics[width=\linewidth]{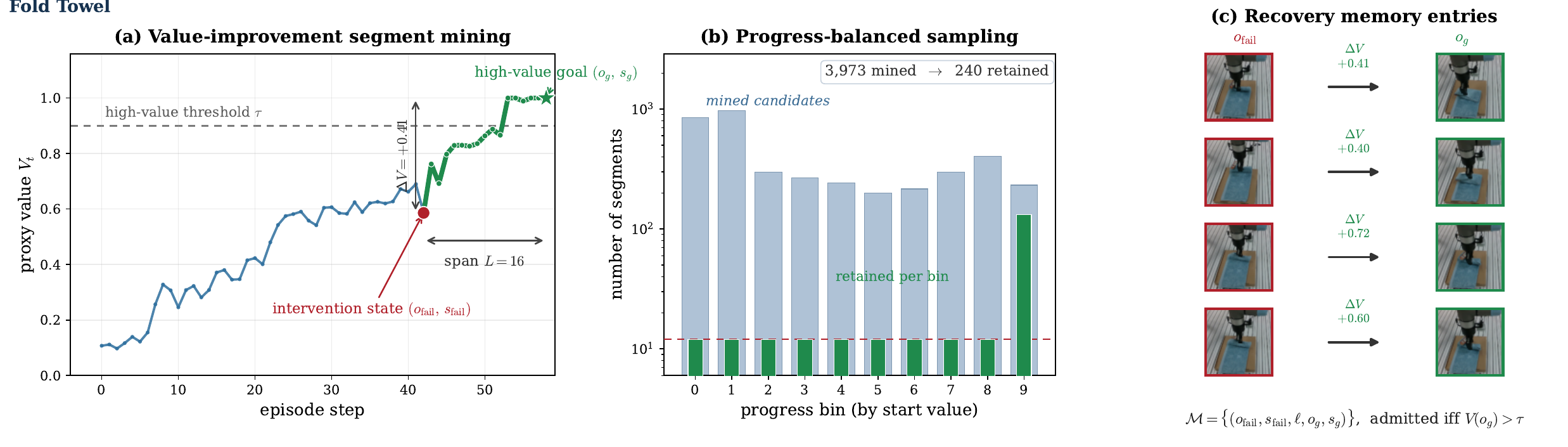}
  \caption{Recovery memory construction for Fold Towel. The longer span $L=16$ matches the
  slower value growth of this deformable, long-horizon task (a); $3{,}973$ mined candidates
  are balanced across progress bins into $240$ retained targets, with the final bin absorbing
  the redistributed remainder (b); and each entry pairs an intervention frame with its
  high-value goal (c).}
  \label{fig:bank_fold}
\end{figure}

At deployment a per-task router maps the current instruction to the matching memory, after
which retrieval operates within that memory. Figure~\ref{fig:bank_fold} illustrates the
construction process on the RAM Insertion data.

\subsection{Target Retrieval}
\label{appendix:retrieval}
Retrieval matches the current context embedding $\phi(o_t,\ell)$ against the memory keys by
cosine similarity and returns the goal of the top-ranked entry. We audited retrieval on the
training rollouts of every task. The top-1 match has a mean cosine similarity above $0.95$
in all tasks, reaching $0.99$ on the insertion and pick tasks, and the retrieved goal lies
on average $0.34$ to $0.56$ in normalized value above the query state. The retrieved goals
are therefore both semantically close to the failure context and substantially higher in
value, which is the property the recovery policy needs.

\begin{table}[h]
\centering
\caption{Retrieval audit on training rollouts. Top-1 similarity is the mean cosine
similarity of the retrieved memory entry; value lift is the mean normalized-value gap
between the retrieved goal and the query state.}
\label{tab:retrieval}
\begin{tabular}{lccc}
\toprule
\textbf{Task} & \textbf{Queries} & \textbf{Top-1 similarity} & \textbf{Value lift} \\
\midrule
Insertion (RAM + Tube) & $14{,}950$ & $0.992$ & $0.452$ \\
Pick Eggplant          & $\phantom{0}5{,}580$  & $0.992$ & $0.563$ \\
Wipe Whiteboard        & $\phantom{0}5{,}529$  & $0.993$ & $0.365$ \\
Fold Towel             & $\phantom{0}2{,}360$  & $0.957$ & $0.339$ \\
\bottomrule
\end{tabular}
\end{table}

\subsection{Goal-conditioned Recovery Policy}
\label{appendix:rechead}
Conditioned on the retrieved goal $g_t$, the current observation, and the instruction, the
recovery policy emits a corrective action chunk of horizon $H=8$ over a $7$-dimensional
end-effector action. Following Eq.~8 of the main paper, the chunk is represented with a FAST
tokenizer as discrete frequency-domain tokens and produced by per-token classification. The
decoder is a compact causal head that reads a context vector pooled from the backbone hidden
state, and it is attached through a shared-residual LoRA adapter of rank $16$ and scaling
$\alpha=32$ so that it reuses the same representation as the intervention module rather than
introducing a separate encoder. Training is behavior cloning on the rollout segments that
connect each mined intervention state to its recovery goal. Because the goal is supplied by
the memory, the otherwise ill-posed problem of choosing where to recover becomes a
goal-reaching problem of how to get there.

\section{Training Procedure and Hyperparameters}
\label{appendix:training}

\subsection{End-to-end Loop}
Algorithm~\ref{alg:uniintervene} shows how \method\ plugs into the online human-in-the-loop
loop. At each step the intervention module runs a single backbone forward pass, predicts the
future latent, estimates the action-conditioned value, and updates the value history. The
risk head turns the recent value trend into a trigger score. When the score crosses the
threshold the module retrieves a high-value goal and overrides the policy action with a
recovery chunk, contributing the resulting transitions to the replay buffer. Human takeover
remains available for the residual cases that the learned recovery does not resolve.

\begin{algorithm}[t]
\caption{\method: agentic intervention in the real-world HiL-RL loop}
\label{alg:uniintervene}
\begin{algorithmic}[1]
\Require policy $\pi_\theta$; intervention module $I_\psi$ with future head $f_{\text{fut}}$,
twin value head $f_Q$, value-history encoder $f_{\text{hist}}$, risk head $f_{\text{risk}}$;
recovery policy $\pi_{\text{rec}}$; memory $\mathcal{M}$; threshold $\tau_{\text{int}}$;
window $K$; horizon $H$; replay buffer $\mathcal{B}$
\For{each training episode}
  \State value history $\mathcal{V} \gets \emptyset$;\quad $t \gets 0$
  \While{episode not done}
    \State observe $o_t$, instruction $\ell$; sample $a_t \sim \pi_\theta(o_t,\ell)$
    \State $h_t \gets f_{\text{vlm}}(o_t,\ell,a_t)$ \Comment{single prompt-only pass}
    \State $\hat z_{t+1} \gets f_{\text{fut}}(h_t)$;\quad $\hat q_t \gets \min f_Q(\hat z_{t+1})$
    \State append current proxy value $V_t$ to $\mathcal{V}$
    \State $\hat R_t \gets f_{\text{risk}}\!\bigl(\hat z_{t+1}, f_{\text{hist}}(\mathcal{V}_{t-K:t})\bigr)$;\quad
           $s^{\text{int}}_t \gets \sigma(\hat R_t)$
    \If{$s^{\text{int}}_t \ge \tau_{\text{int}}$} \Comment{value trend stagnating}
      \State $g_t \gets \arg\max_{j}\ \mathrm{sim}\!\bigl(\phi(o_t,\ell), \phi(o^{\text{fail}}_j,\ell_j)\bigr)$
      \State $A^{\text{rec}}_t \gets \pi_{\text{rec}}(o_t, g_t, \ell)$ \Comment{$H$-step corrective chunk}
      \State execute $A^{\text{rec}}_t$; add recovery transitions to $\mathcal{B}$
    \Else
      \State execute $a_t$; add transition to $\mathcal{B}$
    \EndIf
    \If{human flags unsafe / out-of-distribution}
      \State human takeover; add corrective transitions to $\mathcal{B}$
    \EndIf
    \State $t \gets t + 1$
  \EndWhile
  \State update $\pi_\theta$ from $\mathcal{B}$ with off-policy RL
\EndFor
\end{algorithmic}
\end{algorithm}

\subsection{Component Order and Optimization}
The proxy value function is trained first and then frozen. The intervention module is
trained on rollouts scored by the frozen value function, using the mined trend labels for
the trigger head and the V-JEPA2 targets for the future head. The recovery memory is built
from the same scored rollouts, and the recovery policy is trained by behavior cloning on the
mined segments. All modules use AdamW with a backbone learning rate of $2\times10^{-5}$ and
a head learning rate of $1\times10^{-4}$, run in bf16 with gradient checkpointing enabled
for the larger batch configurations.

\section{Baseline Implementation Details}
\label{appendix:baselines}

\textbf{$\pi_{0.5}$ (SFT).} The imitation baseline fine-tunes a $\pi_{0.5}$ policy with
supervised behavior cloning on $20$ demonstrations per task. It uses the same observation
space and action space as the other methods and serves as the offline starting point on top
of which the online methods improve.

\textbf{HIL-SERL.} The human-in-the-loop RL baseline trains the manipulation policy with
corrective human interventions and off-policy RL updates. It uses the same SpaceMouse
interface and reward definition as \method, so the difference in intervention rate reflects
who decides when to intervene rather than the interface.

\textbf{Failure-aware RL (FA-RL).} The recovery baseline pairs a learned failure predictor
with a recovery policy. Following its design, the failure predictor is a ResNet-18 image
classifier over a two-camera, two-frame stack at $224\times224$, augmented with a value
window of length $8$ projected to $32$ dimensions, and the recovery policy is an
EfficientNet-B0 regressor producing an $8$-step chunk over the $7$-dimensional action with a
Huber loss. Because its trigger is tied to discrete failure events, the failure predictor is
unreliable when failures are visually subtle. On our validation splits its intervention F1
stays low, for example $0.33$ on RAM Insertion and $0.25$ on Fold Towel, which matches the
reduced success and persistent intervention reported in Table~1 of the main paper. The
baseline is trained in a multitask setting with an $80/10/10$ train/validation/test split, a
recovery horizon of $40$ steps, and a recovery bonus of $0.2$, matching the configuration of
the other methods on the shared platform.

\section{Learned Intervention and Recovery Behavior}
\label{appendix:results}

\begin{figure}[h]
  \centering
  \includegraphics[width=\linewidth]{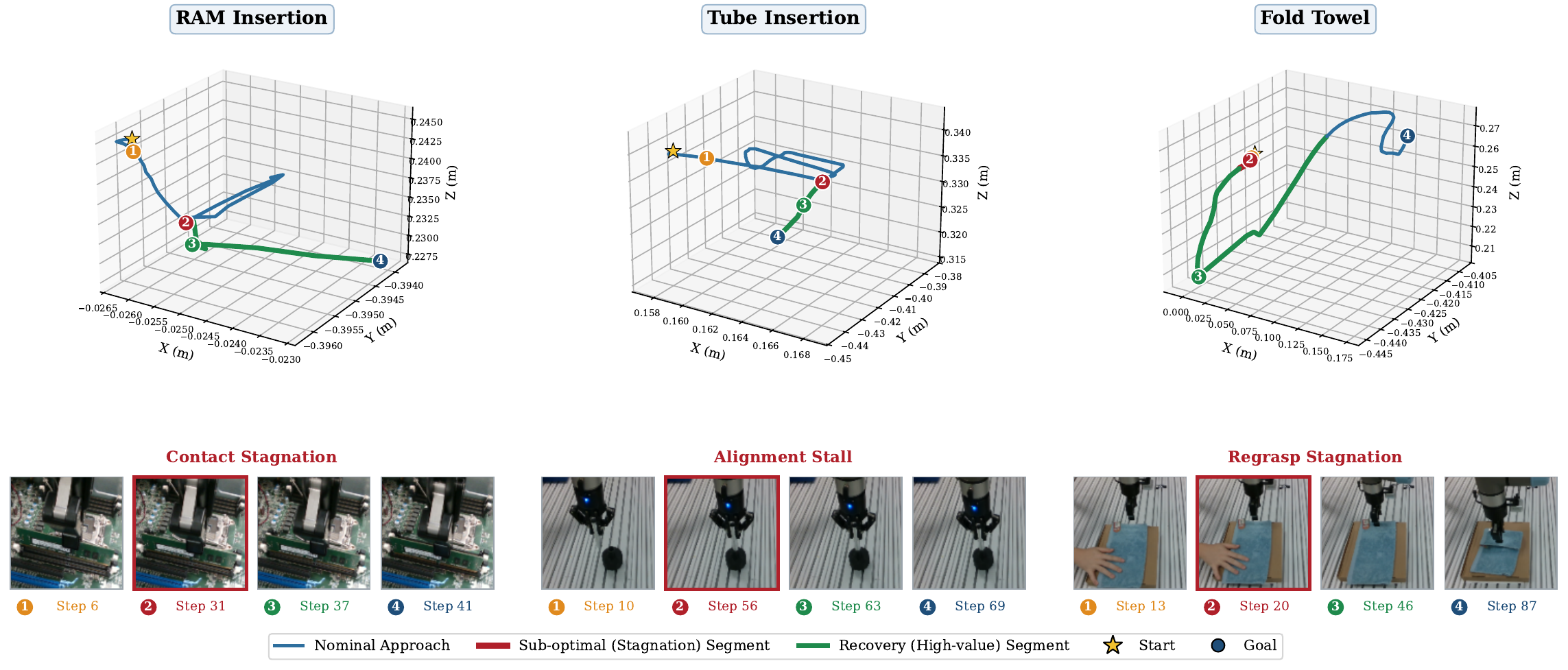}
  \caption{Recorded end-effector trajectories in three dimensions for three tasks, split
  into three phases. The nominal approach is in blue, the sub-optimal stagnation segment
  where end-effector speed collapses is in dark red, and the high-value recovery segment
  along which the policy regains progress is in green. The start is a yellow star and the goal is a blue
  dot. Four steps are marked with numbered circles, and the third-person camera frames below
  each plot show the scene at those steps, with the stagnation step boxed in red. The red
  segment is the low-value stagnation that the temporal value-risk trigger detects, and the
  green segment is the high-value trajectory that recovery is meant to reproduce.}
  \label{fig:traj3d}
\end{figure}

To examine the behavior behind the aggregate metrics, we visualize recorded rollouts in
end-effector space. Figure~\ref{fig:traj3d} shows the three-dimensional path of the gripper
for three tasks, with the stagnation segment highlighted. The stagnation segment is the
contiguous region of lowest end-effector speed away from the endpoints, which is the
signature that the temporal value-risk trigger is built to detect. In each case the path
enters a slow, low-progress region during contact or regrasping, the kind of unproductive
exploration that would otherwise prompt a human takeover, and then resumes toward the goal
along the green recovery segment, which is the high-value trajectory the recovery policy is
trained to reproduce. The third-person camera frames below each plot show the scene at the
four annotated steps,
with the stagnation step boxed in red. The behavior is consistent across the contact-rich
insertions and the deformable Fold Towel task despite their different horizons, which
supports the claim that a single trigger and recovery mechanism transfers across manipulation
types.

\end{document}